\documentclass{pfia}
\usepackage{xcolor}
\usepackage{multirow}
\usepackage{graphicx}
\usepackage{bm}
\usepackage{tabularx}
\usepackage{booktabs}
\usepackage{float}
\usepackage{tablefootnote}
\usepackage{supertabular}

\definecolor{afiablue}{RGB}{61,159,207}
\definecolor{afiared}{RGB}{167,75,68}
\definecolor{afialightblue}{RGB}{158,193,232}

\usepackage{bigstrut}

\newcommand{\field}[1]{\textcolor{afiablue}{#1}}
\newcommand{\val}[1]{\textsf{\textcolor{afialightblue}{#1}}}

\newenvironment{SUBENVcomment}[2]{\color{#1}[#2:~}{]\color{black}}

\definecolor{frederic}{HTML}{5862ed}

\title{\textbf{Large Language Models and Algorithm Execution: Application to an Arithmetic Function}}

\author{Farah Ben Slama, Frédéric Armetta\\
Université Claude Bernard Lyon 1, CNRS,\\ Ecole Centrale de Lyon, INSA Lyon,\\ Université Lumière Lyon 2, LIRIS, UMR5205,\\ 69622 Villeurbanne, France\\
{\texttt{farah.ben-slama@etu.univ-lyon1.fr, frederic.armetta@univ-lyon1.fr}}
}

\date{Janvier 2026}

\begin{document}

\maketitle

\begin{abstract}
Large Language Models (LLMs) have recently developed new advanced functionalities. Their effectiveness relies on statistical learning and generalization capabilities. However, they face limitations in internalizing the data they process and struggle, for instance, to autonomously execute algorithms. In this paper, we investigate the possibility of extending these models' capabilities to algorithm execution through specialized supervised training focused on reasoning decomposition. We introduce a training model called  LLM-DAL (Large Language Model - Decompositional Algorithmic Learning), through which we demonstrate that LLMs' ability to perform complex algorithmic inferences and generalize can be significantly improved when the training method is properly designed to guide the model in its learning process.
\end{abstract}


\section{Introduction}

Deep learning has revolutionized the field of Artificial Intelligence (AI), opening new perspectives in diverse domains such as image recognition, speech recognition, and natural language processing. Thanks to its ability to model complex data structures, it has established itself as an essential tool for addressing increasingly varied problems. However, beyond these well-established applications, one challenge remains largely unexplored: the use of neural networks for learning and executing algorithms.

Neural networks are recognized for their learning and generalization capabilities. Yet, their application to algorithmic learning introduces specific challenges. Indeed, the objective is not limited to classifying data or predicting trends, but rather to enable the network to output the result of an algorithm given input parameters. This approach could automate the execution of complex algorithms without relying on explicit implementation, thereby paving the way for models capable of reasoning in a more systematic and explainable manner, while also generalizing beyond their training data.

The problem raised by Searle, illustrated by the Chinese room thought experiment~\cite{searle_minds_1980}, in which he imagines a person who does not understand Chinese but follows instructions to manipulate Chinese symbols coherently, highlights a fundamental limitation of artificial intelligence systems: although the exchange may appear meaningful to an external observer, the person inside understands nothing of the content. This experiment illustrates that simulating understanding does not correspond to genuine appropriation of the manipulated concepts. Similarly, AI systems manipulate symbols and apply rules without true understanding or appropriation of knowledge. In the case of large language models (LLMs), this limitation is reflected in a lack of knowledge grounding, since their knowledge remains exogenous to the system and does not result from an iterative, subjective construction based on experience.

Thus, in this article, although we do not claim to overcome this fundamental issue of knowledge grounding, we propose a learning method designed to help the system better internalize the resolution of algorithmic problems.

Such learning requires neural networks to capture very long dependencies during algorithmic reasoning, to manipulate variables in their internal representations, and to apply complex transformations to produce the expected output.

A pioneering attempt to tackle this challenge was introduced with the Neural Turing Machine (NTM)~\cite{graves2014neural}, an innovative architecture designed to simulate algorithmic logic through neural networks. While promising, this solution revealed significant limitations, including convergence issues and extreme sensitivity to parameterization~\cite{collier2018implementing}.

Thus, even though LLMs generate linguistically coherent responses, they are often described as “stochastic parrots,” meaning that they reproduce learned information without understanding its logic, especially when it comes to algorithmic reasoning, which restricts their capabilities. This “memorization” of sequences can yield satisfactory results in previously encountered contexts, but does not guarantee that these models can solve complex problems or deeply understand the underlying relationships between concepts.

For instance, in the article published by Razeghi et al.~\cite{razeghi2022impact}, it is shown that performance on arithmetic operations correlates with the frequency of operands in the training corpus. This finding demonstrates that the neural network does not master the underlying logic of the operations but is instead limited to reproducing memorized sequences.

This raises the question of whether such models, based on neural networks, can move beyond this limitation of being “stochastic parrots” and achieve tasks requiring more complex algorithmic reasoning and genuine logical understanding.

We first present a state of the art on language models in Section~\ref{soa}, then describe our proposed model in Section~\ref{modele}. We next detail the experimental setup and obtained results in Section~\ref{experimentations}. Finally, we conclude with a summary of our contributions and future perspectives in Section~\ref{conclusion}.

\section{State of the Art}
\label{soa}

Large Language Models (LLMs) continue to revolutionize the field of artificial intelligence, particularly for complex tasks requiring mathematical reasoning, programming, or scientific applications. In recent years, several well-developed models have emerged, each bringing significant innovations in terms of architecture or performance. The enthusiasm surrounding the dissemination of language models is largely driven by the major AI players. These companies have the financial resources, massive computing infrastructures (GPU, TPU), and research teams required to train and fine-tune such models on enormous datasets.

From a technical standpoint, LLMs rely on deep neural network architectures, thereby inheriting both their capabilities and limitations. Their functioning consists of predicting tokens sequentially, word by word, based on the context formed by preceding words. This corresponds to the memorization of learned \textit{patterns} and statistical generalization, rather than genuine understanding or reasoning. Consequently, their ability to “reason” remains limited to the probabilistic repetition of word sequences, hence the hypothesis of the “stochastic parrot.”

Thus, in this state of the art, we show in Section~\ref{soa1} that significant efforts have been undertaken to design models explicitly oriented towards logical reasoning, with the aim of improving their capacity to solve complex tasks. Section~\ref{soa2} focuses more specifically on mathematical reasoning, and in particular on multiplication, an operation representative of the challenges faced by neural networks.

\subsection{Recent Advances in LLMs Oriented Towards \mbox{Algorithmic Reasoning}}
\label{soa1}

Recent developments in the field of large language models reflect a growing interest in their ability to solve tasks involving algorithmic reasoning, particularly arithmetic and logical tasks. Increasingly, models proposed by major companies and AI research laboratories explicitly integrate these dimensions, underlining the strategic importance assigned to this area of competence for improving the overall performance of LLMs.

In this section, we review several advances in the development of models specialized in this type of reasoning. Table~\hyperref[tab:table0]{1} provides an overview of some recent models such as LLEMMA, Mathstral 7B, QwQ-32B-Preview, as well as OpenAI’s o1 and o3 models. Each model is described according to its specific features, its performance on relevant benchmarks related to its field of expertise, and its main limitations, identified through both feedback from the scientific community and our own experiments.

\begin{table*}[!h]
	\small
	\makebox[\textwidth][c]{%
		
	%\hspace*{-0.5cm}%	
	%\centering
	
	\begin{tabular}{|p{2cm} p{2cm} p{4cm} p{5cm} p{5cm}|}
		\hline
		\textbf{Model} & \textbf{Parameters} & \textbf{Specialization} & \textbf{Performance} & \textbf{Limitations} \\
		
		\hline
		
		\textbf{LLEMMA}\textsuperscript{\href{https://arxiv.org/abs/2310.02961}{a}} & 7B / 34B & Mathematical reasoning, theorem proving & Outperforms Google’s Minerva model\textsuperscript{\href{https://arxiv.org/abs/2206.14858}{f}} on the MATH benchmark\textsuperscript{\href{https://arxiv.org/abs/2103.03874}{g}}, a test dataset evaluating the mathematical reasoning capabilities of language models. & Faces challenges in advanced algebraic reasoning, particularly in proving complex theorems that require deep understanding and sequential reasoning.\\

		\textbf{Mathstral 7B}\textsuperscript{\href{https://mistral.ai/fr/news/mathstral/}{b}} & 7B & Advanced mathematical reasoning & On the MATH benchmark\textsuperscript{\href{https://arxiv.org/abs/2103.03874}{g}}, achieves 56.6\% in direct generation, 68.37\% with \textit{majority voting} (multiple responses generated, selecting the most frequent), and 74.59\% with a \textit{reward model} (selection of the best answer via a reward model). Achieves 63.47\% on the MMLU benchmark\textsuperscript{\href{https://arxiv.org/abs/2009.03300}{h}} evaluating multitask understanding across diverse academic domains. & Struggles with highly complex problems and contexts requiring extended logical reasoning, particularly in handling nonlinear relationships between problem elements.\\
		
		\textbf{QwQ-32B-Preview}\textsuperscript{\href{https://www.reuters.com/technology/alibaba-shares-surge-after-it-unveils-reasoning-model-2025-03-06/}{c}} & 32.5B & Mathematical and algorithmic reasoning & On the MATH-500 benchmark\textsuperscript{\href{https://jurnals.net/qwq-32b-a-new-competitive-reasoning-model/}{i}} and AIME\textsuperscript{\href{https://www.deeplearning.ai/the-batch/qwens-qwq-32b-preview-packs-a-big-punch/}{j}}, the model achieves success rates of 90.6\% and 50\% respectively. It also reaches 65.2\% on GPQA\textsuperscript{\href{https://www.datacamp.com/blog/qwq-32b-preview}{k}} and 50\% on LiveCodeBench\textsuperscript{\href{https://jurnals.net/qwq-32b-a-new-competitive-reasoning-model/}{l}}.& Shows notable limitations: language mixing, recursive reasoning loops, limited commonsense understanding, and ethical and safety concerns\textsuperscript{\href{https://qwenlm.github.io/blog/qwq-32b-preview/}{m}}.\\
		
		\textbf{OpenAI o1}\textsuperscript{\href{https://fr.wikipedia.org/wiki/OpenAI_o3}{d}} & Not specified & Complex reasoning, RLHF (Reinforcement Learning from Human Feedback) & Estimated IQ of 120 based on Mensa equivalences\textsuperscript{\href{https://www.mensa.org/}{a}}, ranks in the top 500 of the AIME benchmark\textsuperscript{\href{https://www.aime.org/}{b}}, and achieves the 89th percentile on Codeforces\textsuperscript{\href{https://codeforces.com/}{c}}, a competitive programming platform. Also shows strong performance on the GPQA benchmark\textsuperscript{\href{https://www.gpqa.com/}{d}}, measuring factual and conceptual reasoning. & Shows weaknesses in long, multi-step reasoning, especially in open contexts or when implicit inferences are required\textsuperscript{\href{https://www.openai.com/blog/chatgpt-4}{e}}. \\
		
		\textbf{OpenAI o3}\textsuperscript{\href{https://fr.wikipedia.org/wiki/OpenAI_o3}{e}} & Not specified & Chain-of-thought, algorithmic reasoning & On the GPQA Diamond benchmark\textsuperscript{\href{https://www.gpqa.com/}{f}}, o3 achieves 87.7\%. It surpasses its predecessor o1 by a factor of 3 on the ARC-AGI test\textsuperscript{\href{https://www.openai.com/research/}{g}}, which evaluates analogical generalization capabilities. It also reaches 71.7\% on SWE-bench Verified\textsuperscript{\href{https://www.swe-bench.com/}{h}}, measuring the ability to resolve software engineering problems based on real bug reports, and achieves a rating of 2727 Elo on Codeforces\textsuperscript{\href{https://codeforces.com/}{i}}. & Still struggles to maintain perfect coherence in very long or uncertain chains of reasoning, and may fail to effectively synthesize multiple sources of information in unseen scenarios\textsuperscript{\href{https://www.openai.com/research/}{j}}. \\
		
		\hline

	\end{tabular}

	}
	\caption{\normalsize{Recent reasoning-oriented models evaluated in various ways according to their specialization}}

	\textsuperscript{a}~\href{https://arxiv.org/abs/2310.02961}{arXiv – Llemma}
	\textsuperscript{b}~\href{https://mistral.ai/fr/news/mathstral/}{Mistral – Mathstral} \quad
	\textsuperscript{c}~\href{https://www.reuters.com/technology/alibaba-shares-surge-after-it-unveils-reasoning-model-2025-03-06/}{Reuters – QwQ} \quad
	\textsuperscript{d}~\href{https://fr.wikipedia.org/wiki/OpenAI_o3}{Wikipedia – OpenAI o1} \quad
	\textsuperscript{e}~\href{https://fr.wikipedia.org/wiki/OpenAI_o3}{Wikipedia – OpenAI o3} \quad
	\textsuperscript{f}~\href{https://arxiv.org/abs/2206.14858}{arXiv – Minerva} \quad
	\textsuperscript{g}~\href{https://arxiv.org/abs/2103.03874}{arXiv – MATH benchmark} \quad
	\textsuperscript{h}~\href{https://arxiv.org/abs/2009.03300}{arXiv – MMLU} \quad
	\textsuperscript{i}~\href{https://jurnals.net/qwq-32b-a-new-competitive-reasoning-model/}{Jurnals.net – QwQ benchmark} \quad
	\textsuperscript{j}~\href{https://www.deeplearning.ai/the-batch/qwens-qwq-32b-preview-packs-a-big-punch/}{DeepLearning.ai – QwQ AIME} \quad
	\textsuperscript{k}~\href{https://www.datacamp.com/blog/qwq-32b-preview}{DataCamp – QwQ GPQA} \quad
	\textsuperscript{l}~\href{https://jurnals.net/qwq-32b-a-new-competitive-reasoning-model/}{Jurnals.net – LiveCodeBench} \quad
	\textsuperscript{m}~\href{https://qwenlm.github.io/blog/qwq-32b-preview/}{QwenLM blog – limitations QwQ} \quad
	\textsuperscript{n}~\href{https://www.mensa.org/}{Mensa – QI estimé} \quad
	\textsuperscript{o}~\href{https://www.aime.org/}{AIME – benchmark math} \quad
	\textsuperscript{p}~\href{https://codeforces.com/}{Codeforces – compétition algorithmique} \quad
	\textsuperscript{q}~\href{https://www.gpqa.com/}{GPQA benchmark} \quad
	\textsuperscript{r}~\href{https://www.openai.com/blog/chatgpt-4}{OpenAI blog – limitations o1} \quad
	\textsuperscript{s}~\href{https://www.openai.com/research/}{OpenAI Research – benchmarks o3} \quad
	\textsuperscript{t}~\href{https://www.swe-bench.com/}{SWE-bench – software engineering benchmark}
	
	\label{tab:table0}

\end{table*}

\normalsize

Table \hyperref[tab:table0]{1} clearly illustrates the growing interest within both the scientific and industrial communities in developing language models capable of algorithmic reasoning. This trend is reflected in the increasing number of studies aimed at equipping LLMs with advanced cognitive mechanisms that enable them to tackle complex tasks, such as mathematical problems or structured logical processes. These efforts highlight a strong desire to overcome the limitations of current approaches, which rely predominantly on statistical prediction, by incorporating abilities that more closely resemble human reasoning—abilities that are essential for solving tasks requiring step-by-step execution or algorithmic generalization.

\subsection{Limits of Neural Networks in Algorithmic Learning: the Case of Multiplication}
\label{soa2}

In addition to recent advances in language models capable of handling reasoning tasks, notable efforts have been made to endow neural architectures with a form of procedural reasoning; either by extending standard architectures or by hybridizing neural models with external tools that can execute the task and return the result to the network without the network performing the reasoning itself. However, despite these advances, neural networks still suffer from structural limitations that compromise their ability to learn and execute algorithms reliably and in a generalizable way.

In our study, we tested a commonly used approach that delegates the execution of the multiplication algorithm to external \textit{APIs} capable of performing complex mathematical computations and returning the result to the language model. In this setup, the model has a predefined list of \textit{APIs} it can call to execute the algorithmic task outside its architecture, then retrieve the result and integrate it into its response.

This method proves effective for simple algorithmic tasks—often achieving a 100\% success rate—but it has drawbacks, including difficulty handling nested calls, dependency on external services, API latency, and data-privacy concerns. Moreover, it does not enable the language model to learn to perform the algorithm itself, which limits its ability to generalize beyond the set of external resources available to it.

The multiplication problem used to illustrate our learning model, though apparently simple, is nonetheless difficult when solved by a neural network or an LLM. A study by Hoshen et al. \cite{Hoshen:2016:VLA:3016387.3016429} shows that multiplication is very hard to learn \emph{end-to-end} using a multilayer perceptron.

For large language models, performance on arithmetic operations is, for instance, correlated with the frequency of the operands in the training corpus \cite{razeghi2022impact}. This indicates that the neural network does not master the underlying logic of the operations but is limited to reproducing memorized sequences. The result supports the legitimacy of the “stochastic parrot” hypothesis: these models merely repeat patterns observed in the data without genuine understanding or the capacity to generalize beyond those specific examples.

Algorithmic inference requires step-by-step execution with structured updates to intermediate data, which exceeds the capabilities of LLMs in their common usage—models that primarily operate by pattern association and probabilistic text generation. Unlike architectures designed for symbolic execution, LLMs lack a persistent working memory and an internal mechanism to apply rules iteratively under a controlled state, which limits their aptitude for performing complex algorithmic reasoning.

One strategy to facilitate learning of complex tasks is to scaffold the model so it learns tasks of increasing complexity \cite{bengio2009curriculum}. Such decomposition has improved end-to-end learning of multiplication for recurrent neural networks \cite{armetta2023apprentissage, armetta2023algorithmic}. Despite these promising results, these works remain constrained by the intrinsic difficulty of algorithmic learning—particularly long-range dependencies and trainability problems. The language models used struggle to generalize when input variability is large. Moreover, these methods still rely on full supervision, raising the question of the model’s ability to infer missing intermediate steps.

In this paper, we therefore aim to teach a large language model to perform complex inferences. We draw inspiration from Chain of Thought (CoT) reasoning \cite{wei2022chain}, which improves LLMs’ ability to structure their reasoning, by incorporating tasks learned with progressively increasing complexity into the reasoning process and promoting algorithmic transfer between them.

\section{Learning Framework}
\label{modele}

In this section, we detail the learning methodology used to train a large language model on the multiplication task, along with the training corpus we constructed.

The core idea of our approach, introduced under the name  LLM-DAL (Large Language Model – Decompositional Algorithmic Learning) , is based on breaking down the studied task into several simpler subtasks, each representing a key step of the overall process. This strategy can be applied to many algorithmic problems, and here we adapt it specifically to multiplication. The order in which the subtasks are learned is carefully structured to strengthen the model’s ability to generalize its knowledge and to learn progressively.

The interest of this method is twofold: it not only improves the model’s reasoning and generalization abilities but also reduces task complexity by splitting it into digestible and more easily learnable steps.

Thus, the fine-tuning approach employed for this learning model has a structure that could be applied to other algorithmic tasks, and it demonstrates how progressive learning—structured into subtasks—can enhance a language model’s ability to reason and solve complex problems.

\subsection{\textit{Chain of Thought} and Task Decomposition}

Chain of Thought (CoT) reasoning \cite{wei2022chain} is an approach that consists of breaking down a complex task into several simpler subtasks, which are then solved sequentially to reach the final result. This method simplifies the overall task by dividing it into intermediate steps, while enabling the language model to reason more systematically and generalize more effectively.

In the context of learning multiplication, this approach allows us to decompose the operation into elementary subtasks, which are then combined to obtain the final result.

These subtasks include:
\begin{itemize}
\item  Digit-by-digit multiplication, $t1_{mult}$ : multiplying each digit of the multiplier by each digit of the multiplicand. For this, the language model must rely on knowledge of single-digit multiplication tables.
\item  Addition of partial results, $t2_{add}$ : adding the result of the intermediate digit-by-digit multiplication with the carry obtained from the previous multiplication. The additions to be learned therefore involve at most two digits: the product of the digit-by-digit multiplication and the carry, which is always a single digit.
\item  Digit extraction, $t3_{extract}$ : extracting the digit at the $n$-th position of a given number.
\item  Left concatenation of digits, $t4_{concat}$ : adding a digit to the left of a given number.
\end{itemize}

After progressive supervised learning on these subtasks, the trained model will be able to generate a natural-language description of the reasoning it follows to perform the multiplication of two numbers. This description allows us to evaluate the model’s ability to reason logically and adapt to complex examples, even outside its training corpus.

For example, when asked to compute the product of 5847 by 2, the model should be able to generate a step-by-step reasoning description that leads to the expected result:

\textbf{\texttt{multiplying 5847 by 2: 5847*2=}}\\
\texttt{carry=0}\\
\texttt{digit 1 of 5847 is 7}\\
\texttt{multiply digit 1 of 5847 which is 7 by 2: temp\_mult=7*2=14}\\
\texttt{Add the multiplication result to the carry: temp\_add=carry+temp\_mult=0+14=14}\\
\texttt{compare the addition result to 10: temp\_add=14>=10}\\
\texttt{the first digit of temp\_add=14 is fd\_result=4}\\
\texttt{first digit of temp\_add which is 4 is concatenated to the left of temp\_result=}\\
\texttt{the result of the concatenation is 4}\\
\texttt{the second digit of temp\_add is 1 which will be the value of the carry: carry=1}\\

\texttt{carry=1}\\
\texttt{digit 2 of 5847 is 4}\\
\texttt{multiply digit 2 of 5847 which is 4 by 2: temp\_mult=4*2=8}\\
\texttt{Add the multiplication result to the carry: temp\_add=carry+temp\_mult=1+8=9}\\
\texttt{compare the addition result to 10: temp\_add=9<10}\\
\texttt{carry=0}\\
\texttt{temp\_add is concatenated to the left of temporary\_result=4}\\

\texttt{...}

\texttt{final carry=1>0}\\
\texttt{the final carry which is 1 is concatenated to the left of the final result which is 1694}\\
\texttt{the result of the concatenation is 11694}\\
\texttt{the final\_result is 11694}\\
\label{eg1}

\subsection{Training Corpus}

As part of our learning model, training corpora were synthetically generated for each of the identified subtasks. This generation was necessary due to the absence of existing datasets that precisely met our experimental requirements and the level of granularity needed for supervised algorithmic learning. Each generated dataset entry was accompanied by a natural-language description, allowing the language model to master the task more effectively.

Examples of these training data are presented in the following tables:

For task $t1_{mult}$, an input–output example is presented in Table \ref{tab:table1}. Each example consists of the multiplication operation (multiplier, multiplicand, and operators) and the corresponding multiplication result.

\begin{table}[htbp]
\centering
\begin{tabular}{ll}
\hline
Input & 6 by 8: 6*8 =  \\
\hline
Output & 48 \\
\hline
\end{tabular}
\caption{Digit-by-digit multiplication task}
\label{tab:table1}
\end{table}

For task $t2_{add}$, an input–output example is presented in Table \ref{tab:table2}. Each example consists of the addition operation (the two numbers and the operator) and the corresponding sum.

\begin{table}[htbp]
\centering
\begin{tabular}{ll}
\hline
Input & 12 plus 6: 12+6= \\
\hline
Output & 18 \\
\hline
\end{tabular}
\caption{Addition task: a two-digit number plus a single digit}
\label{tab:table2}
\end{table}

For task $t3_{concat}$, an input–output example is presented in Table \ref{tab:table3}. Each example consists of the number to be concatenated on the left and the destination number, along with the result of the concatenation.

\begin{table}[htbp]
\centering
\begin{tabular}{ll}
\hline
Input & Concatenating 16 to 375347 on the left gives \\
\hline
Output & 16375347 \\
\hline
\end{tabular}
\caption{Left concatenation of one number to another}
\label{tab:table3}
\end{table}

For task $t4_{extract}$, an input–output example is presented in Table \ref{tab:table4}. Each example consists of the number and the digit position to be extracted, along with the extracted digit as output.

\begin{table}[htbp]
\centering
\begin{tabular}{ll}
\hline
Input & The 3rd digit from 393721 is \\
\hline
Output & 3 \\
\hline
\end{tabular}
\caption{Digit extraction task}
\label{tab:table4}
\end{table}

Finally, a combined training corpus was created by merging the four subtasks to generate a complete description of the reasoning required to perform the multiplication of two numbers, as illustrated in the example at the end of Section \ref{eg1}. This approach allows the model to consolidate the knowledge acquired from the individual subtasks and produce a detailed explanation of the multiplication process.

The goal of these training procedures is to leverage the text-generation capabilities of LLMs to teach them to carry out a complex algorithmic task by combining algorithms expressed in natural language with synthetically generated training data.

\section{Experiments}
\label{experimentations}
\subsection{Experimental Context}

The LLM used to carry out all learning steps is \textbf{Llama 3.2}, a 3-billion parameter language model in its Instruct version\footnote{\label{llama} \href{https://huggingface.co/meta-llama/Llama-3.2-3B-Instruct}{https://huggingface.co/meta-llama/Llama-3.2-3B-Instruct}}.  

This version has been fine-tuned to follow instructions. Unlike non-instruction-tuned models, which simply continue the prompt, these models are designed to respond precisely and contextually to user queries.  

We selected this model for its suitability to the tasks at hand, the computational resources available, and its strong performance relative to its size on the various benchmarks on which it was evaluated.  

Within our algorithmic framework, the working memory required for manipulating intermediate variables is integrated into the context of the LLM. We leveraged the extended context window of Llama 3.2, which reaches up to 128,000 tokens\textsuperscript{\ref{llama}}. This capacity allows the model to retain and process complex, long-form sequences, which is crucial for learning multi-step arithmetic tasks.  

To ensure reproducibility, we provide the task-specific training corpora we constructed, accessible via the links in the footnotes.  

For task $t1_{mult}$, a training corpus of all possible combinations of two-digit multiplications between 0 and 9 was generated, yielding a total of 100 examples. Each example consists of the multiplication operation (multiplier, multiplicand, and operator) and its result.  

Similarly, for task $t2_{add}$, we generated a training corpus covering all possible combinations of additions with digits from 0 to 9, yielding 1000 examples. Each example consists of the addition operation (the two numbers and the operator) and its result.  

The two corpora corresponding to $t1_{mult}$ and $t2_{add}$ were merged into a single dataset of 1100 examples of addition and multiplication\footnote{\href{https://huggingface.co/datasets/farahbs/onetwo-digits-mult-add}{https://huggingface.co/datasets/farahbs/onetwo-digits-mult-add}}, which was then used for the initial model training stage.  

For task $t3_{extract}$, we generated a training corpus of 5000 random examples of digit extraction from numbers of 4 to 6 digits\footnote{\href{https://huggingface.co/datasets/farahbs/extraction\_db}{https://huggingface.co/datasets/farahbs/extraction\_db}}. Each example consists of the number, the digit position to extract, and the extracted digit.  

For task $t4_{concat}$, we generated a training corpus of 5000 random examples of concatenating a 1- or 2-digit number to the left of a 4- to 6-digit number\footnote{\href{https://huggingface.co/datasets/farahbs/left-concat}{https://huggingface.co/datasets/farahbs/left-concat}}. Each example consists of the left-side number, the target number, and the concatenation result.  

All the above datasets were split as follows: 70\%, i.e., 3500 examples, for training and 30\%, i.e., 1500 examples, for evaluation.  

Finally, we generated an additional dataset using an algorithm that combines the four sub-tasks to produce reasoning chains for performing the overall multiplication of two numbers\footnote{\href{https://huggingface.co/datasets/farahbs/cot-multiplication-2k}{https://huggingface.co/datasets/farahbs/cot-multiplication-2k}}. This dataset contains 2000 random examples of multiplications between a 4- to 6-digit number and a single-digit number. Each example consists of the two numbers to multiply, the reasoning description, and the multiplication result. The dataset is relatively small compared to the others due to the length and complexity of the generated reasoning descriptions, which make training time-consuming. It was split as follows: 90\%, i.e., 1800 examples, for training and 10\%, i.e., 200 examples, for evaluation. This split differs from the other datasets because of the dataset’s relatively small size and the complexity of the reasoning descriptions to be learned by the model.

\subsection{Incremental Learning Process}

The experimental protocol was designed in several intermediate training stages, allowing the model to gradually acquire the various sub-tasks required to perform a complete multiplication. This progressive approach, detailed in Table \ref{tab:table5}, aims to strengthen the model's understanding by providing structured and decomposed learning before tackling the overall task.

\begin{table}[htbp]
  \centering
  \begin{tabular}{lll}
    \hline
    Step & Task & Type of Training \\
    \hline
    1 & $t1_{mult}$ + $t2_{add}$ & overfitting \\
    2 & $t3_{extract}$ + $t4_{concat}$ & \textit{fine-tuning} \\
    3 & $t_{globale}$ & \textit{fine-tuning} \\
    \hline
  \end{tabular}
  \caption{Training configurations for algorithmic learning of multiplication}
  \label{tab:table5}
\end{table}

The first stage consisted of \textit{fine-tuning} the base Llama-3.2-3b-instruct model on tasks $t1_{mult}$ and $t2_{add}$, resulting in model M1. Given the critical importance of these two sub-tasks relative to the others, we opted for overfitting specifically on these tasks. Although limited in number, these tasks are fundamental, as they form the basic operations on which the model relies to perform complete and coherent reasoning. Generalization will occur during the execution of the full multiplication operation.

The combined dataset was used solely for training, with a validation split but without evaluation on unseen examples. Fine-tuning was stopped once the loss function plateaued, indicating that the model had satisfactorily learned the addition and multiplication tasks.

The next stage was to fine-tune model M1 on task $t3_{extract}$, resulting in model M2. This task is crucial for the overall process, as it enables the extraction of digits needed to perform digit-by-digit multiplication.

Subsequently, model M2 was fine-tuned on task $t4_{concat}$, yielding model M3. This task is also critical for the full reasoning process, as it allows the combination of extracted digits and intermediate results to produce the final multiplication output.

Models M2 and M3 were trained on their respective training corpora using the same fine-tuning approach as for M1. The progression of the loss function was monitored throughout training on both the training and validation sets to avoid overfitting (where the model memorizes the training data without generalizing) or underfitting (where it fails to extract useful patterns from the data). Once fine-tuning was complete, the models’ performance was evaluated on the evaluation set to measure their ability to generalize to unseen examples.

Finally, the last stage involved training model M3 on the global multiplication task dataset, resulting in the final LLM-DAL model.

After this supervised learning on the global multiplication task, the model’s performance was evaluated on the evaluation set. However, during initial evaluations, the model struggled to produce a complete response that included all intermediate steps necessary for reasoning through the multiplication. To address this, a recursive \textit{prompting} mechanism was implemented to allow the model to generate a complete multi-step response. This mechanism consists of asking the model to generate a partial response to the posed question, then requesting the continuation based on the previous output. A parser checks whether a sentence indicating the end of the response has been produced; if not, the model is prompted again to continue. This process is repeated until the model produces a full response to the question or a maximum of 10 iterations is reached, preventing divergence or indefinite repetition.

\subsection{Results}

Table \ref{tab:table6} below illustrates the evaluation results of the model trained on the global multiplication task using LLM-DAL. This approach allowed the model to be trained on the various identified sub-tasks, explore the final global multiplication task, and leverage recursive \textit{prompting} to generate complete responses for this task. The obtained performance is compared to that of the Vanilla model, corresponding to the initial version of \textit{Llama-3.2-instruct}, trained and evaluated directly on the global task training corpus without intermediate steps, in order to assess the true impact of our training method.

\begin{table}[htbp]
  \centering
  \begin{tabular}{llll}
    \hline
    \multirow{1}{*}{Task} &Vanilla Model & \textbf{LLM-DAL}\\  
    \hline
    \multirow{1}{*}{Accuracy Rate} &13.5\% & \textbf{42.1\%} \\  
    \hline
  \end{tabular}
  \centering \caption{Evaluation results on the global multiplication task}
    \label{tab:table6}
\end{table}

The results show that the model trained using LLM-DAL achieved a performance improvement of over 28\% on reasoning and the global multiplication task.

Table \ref{tab:table7} presents the evaluation results of the initial model trained separately on each sub-task. The results indicate that the model achieved nearly 100\% performance on each intermediate step, which allowed us to proceed to the next stage of training.

\begin{table}[htbp]
  \centering
  \begin{tabular}{llll}
    \hline
    \multirow{1}{*}{Task} & Accuracy Rate  \\  
    \hline
    \multirow{1}{*}{$t1_{mult}$ + $t2_{add}$} & \textbf{100\%} \\  
    {$t3_{extract}$} & \textbf{97.6\%} \\
    {$t4_{concat}$} & \textbf{99.2\%} \\
    \hline
  \end{tabular}
  \centering \caption{Accuracy rate per sub-task during training}
    \label{tab:table7}
\end{table}

For the following experiments, model M1, which was specifically trained on tasks $t1_{mult}$ and $t2_{add}$, was selected as the starting point for subsequent training.

The next training stage consisted of supervised learning of model M1 on task $t3_{extract}$, resulting in model M2. The accuracy results obtained on the training dataset for this task are presented in Table \ref{tab:table8}.

\begin{table}[htbp]
  \centering
  \begin{tabular}{llll}
    \hline
    \multirow{1}{*}{Task} & Accuracy Rate  \\  
    \hline
    \multirow{1}{*}{$t1_{mult}$ + $t2_{add}$} & \textbf{100\%} \\  
    {$t3_{extract}$} & \textbf{99.3\%} \\
    \hline
  \end{tabular}
  \centering \caption{Accuracy rate per sub-task during the second training stage}
    \label{tab:table8}
\end{table}

The final experiment took model M2 as the starting point, which was then trained on task $t4_{concat}$, resulting in the \textbf{M3 model}.

The results of this experiment, shown in Table \ref{tab:table9}, indicate that model M3 achieved the following accuracies across the different tasks:

\begin{table}[htbp]
  \centering
  \begin{tabular}{llll}
    \hline
    \multirow{1}{*}{Task} & Accuracy Rate  \\  
    \hline
    \multirow{1}{*}{$t1_{mult}$ + $t2_{add}$} & \textbf{99.81\%} \\  
    {$t3_{extract}$} & \textbf{99\%} \\
    {$t4_{concat}$} & \textbf{99.53\%} \\
    \hline
  \end{tabular}
  \centering \caption{Accuracy rate per sub-task during combined training}
      \label{tab:table9}
\end{table}

These results demonstrate that fine-tuning on the different tasks improved the model's performance on these specific tasks. The model showed its ability to generalize knowledge and provide accurate and coherent results on tasks it had not seen during training, as well as its ability to transfer knowledge learned from specific tasks to more complex tasks. This knowledge transfer process, where a model uses information learned from one task or dataset to improve performance on another related task or dataset, plays a key role in learning the global multiplication task, which requires mastery of all identified sub-tasks.

Furthermore, our introduced learning method allowed us to operate in a frugal AI setting, where the model was able to acquire the capacity to solve operations logically and adapt to new situations beyond the training data. Due to the small size of the model and training datasets, the model learned to perform a complex algorithmic task in a reasonable time and with limited computing power. For example, it was able to combine the four sub-tasks to generate a natural language description of the reasoning behind the multiplication of two numbers in less than 2 hours on a machine equipped with an NVIDIA A40 GPU with 48 GB of memory. This contrasts with larger models and alternative training methods, which require weeks of training on expensive computing infrastructure.

\section{Conclusion} 
\label{conclusion}

Algorithmic learning is a central challenge to overcome the limitations of LLMs, enable better mastery of the data they manipulate, and move beyond the "stochastic parrot" stereotype, which restricts their capacity to reason autonomously.

Although several works explore this direction, it remains an open and current problem. Integrating structured algorithmic learning could allow language models to generalize better and produce higher-quality results, based more on logical reasoning rather than simple memorization.

In this work, we focused on multiplication, an inherently algorithmic task whose learning proved particularly difficult for LLMs. We highlighted the limitations of current models when faced with this task and proposed an approach that allows LLMs to acquire a resolution logic within reasonable training times. It is important to note, however, that this problem is treated as an illustrative case study, through which we explore the current limits of these models, focusing on their ability to generalize and reason algorithmically.

We assessed the interest and potential benefits of algorithmic learning to improve LLMs’ ability to execute algorithms. To do this, we introduced LLM-DAL, a multi-step learning model combining supervised sub-task learning and structured prompting to reinforce reasoning coherence. The results show a significant improvement in model accuracy as well as a better generalization capacity on similar tasks.

Finally, this work opens several avenues for future research. Applying the approach to other algorithmic tasks would allow the evaluation of the robustness and transferability of the studied methods. Moreover, it would be interesting to explore methods to automatically extract Chain of Thought reasoning without supervision, in order to enhance the models’ ability to structure their learning more autonomously and efficiently.

\bibliographystyle{plain}
\bibliography{biblio-ch-pfia}

\end{document}